\documentclass[twocolumn,10pt]{article}

\usepackage[square,numbers]{natbib}
\usepackage[english]{babel}
\usepackage[utf8]{inputenc} 
\usepackage[T1]{fontenc}    
\usepackage{hyperref}       
\usepackage{url}            
\usepackage{booktabs}       
\usepackage{amsfonts}       
\usepackage{amsmath}
\usepackage{amssymb}
\usepackage{nicefrac}       
\usepackage{microtype}      
\usepackage{lipsum}		    
\usepackage{algorithm}
\usepackage{algpseudocode}
\usepackage{graphicx}
\usepackage{float} 
\usepackage{subfigure} 
\usepackage{multicol}
\usepackage{geometry}
\usepackage{caption}
\usepackage{times}
\usepackage{indentfirst}
\usepackage{titlesec}
\usepackage{threeparttable}

\title{\Large \textbf{Double Prioritized State Recycled Experience Replay\\
		\hspace*{\fill}}}

\author{
  \large
  Fanchen Bu, Dong Eui Chang \\
  \textit{Korea Advanced Institute of Science and Technology}\\
  \href{mailto:boqvezen97@kaist.ac.kr}{\textit{boqvezen97@kaist.ac.kr}},  
  \href{mailto:dechang@kaist.ac.kr}{\textit{dechang@kaist.ac.kr}}\\
}

\date{}

\makeatletter

\newenvironment{breakablealgorithm}
{
		\refstepcounter{algorithm}
		\hrule height2pt depth0pt \kern2pt
		\renewcommand{\caption}[2][\relax]{
			{\raggedright\textbf{\ALG@name~\thealgorithm} ##2\par}%
			\ifx\relax##1\relax 
			\addcontentsline{loa}{algorithm}{\protect\numberline{\thealgorithm}##2}%
			\else 
			\addcontentsline{loa}{algorithm}{\protect\numberline{\thealgorithm}##1}%
			\fi
			\kern2pt\hrule\kern2pt
		}
	}{
		\kern2pt\hrule\relax
}
\makeatother

\begin{document}
\maketitle

\makeatletter
\renewenvironment{abstract}{%
	\if@twocolumn
	\section*{\centering \large \abstractname}%
	\else 
	\begin{center}%
		{\bfseries \Large\abstractname\vspace{\z@}}
	\end{center}%
	\quotation
	\fi}
{\if@twocolumn\else\endquotation\fi}
\makeatother

\begin{abstract}
Experience replay enables online reinforcement learning agents to store and reuse the previous experiences of interacting with the environment. In the original method, the experiences are sampled and replayed uniformly at random. A prior work called prioritized experience replay was developed where experiences are prioritized, so as to replay experiences seeming to be more important more frequently.
In this paper, we develop a method called double-prioritized state-recycled (DPSR) experience replay, prioritizing the experiences in both training stage and storing stage, as well as replacing the experiences in the memory with state recycling to make the best of experiences that seem to have low priorities temporarily.
We used this method in Deep Q-Networks (DQN), and achieved a state-of-the-art result, outperforming the original method and prioritized experience replay on many Atari games.
\end{abstract}

\def\keywordname{{\textbf{Keywords:}}}%
\def\keywords#1{\par\addvspace\medskipamount{\rightskip=0pt plus1cm
		\def\and{\ifhmode\unskip\nobreak\fi\ {, }
		}\noindent\keywordname\enspace\ignorespaces#1\par}}


\titleformat{\section}[block]{\large \bfseries}{\arabic{section}. }{0pt}{}[]
\titleformat{\subsection}[block]{\normalsize \bfseries}{\arabic{section}.\arabic{subsection} }{0pt}{}[]

\section{Introduction}
In online reinforcement learning, agents learn to change the parameters of the policy while interacting with the environment at the same time. Without remembering the previous experiences, agents are only able to update the parameters immediately after each single step, which may affect the efficiency of training process as some experiences can be rare but significant. 

To tackle this issue, \textit{experience replay} [\citealp{lin1992self}] was introduced where the experiences are stored in memory and utilized more methodically. The prominent effect of experience play was proved by its application in Deep Q-Networks (DQN) [\citealp{mnih2013playing, mnih2015human}], on account of its capability to break the temporal correlations of the sequential experiences and palliate the non-stationary distribution problem. Generally, with experience replay, we can downsize the amount of the experiences required for the training process, and therefore reduce the main computational cost in most cases of Reinforcement Learning. 

In the original version of experience replay algorithm, a uniform sampling strategy is used, which can hardly harmonize with the different significance of experiences, and therefore lose some efficiency of learning. Then, \textit{prioritized experience replay} [\citealp{schaul2015prioritized}] was developed to address this issue by directly and simply prioritizing the experiences with higher temporal difference (TD) errors when sampling experiences for training. 

In this paper, we introduce double-prioritized state-recycled (DPSR) experience replay prioritizing the experiences by some standard both in sampling and replacing, as well as executing state recycling, which makes use of some old and likely useless experiences. Our key idea is to keep the experiences that are more useful in the replay buffer for a longer time and make them tend to be sampled more easily and frequently. By keeping a high-quality replay buffer, a Reinforcement Learning agent can waste less time and learn more effectively. 

Specifically, the main contributions of our work are listed as follows: 
\begin{itemize}
	\item [1)] 
	We extended the previous prioritized experience replay and developed a novel experience replay algorithm, double-prioritized experience replay, where the experiences in replay buffer are prioritized in both sampling stage and replacing stage for training the agent.  
	\item [2)]
	We developed state recycling, a special technique to reuse and update the experiences based on old ones, and integrate it with the double-prioritized experience replay algorithm, eventually forming the double-prioritized state-recycled (DPSR) experience replay.    
	\item [3)]
	We applied and tested our DPSR experience replay on Atari games with Deep Q-Networks (DQN). We compared the performance of our method with both the original experience replay and prioritized experience replay. In most Atari games tested, DPSR experience replay outperforms both baseline methods and achieves state-of-the-art results.
\end{itemize}

\section{Background}
\subsection{Problem Statement}
Consider non-discount reinforcement learning (RL), which can be represented by a quadruple ($\mathcal{S, A, P, R}$), where $\mathcal{S}$ is the set of states, $\mathcal{A}$ is the set of actions, $\mathcal{P}$ : $\mathcal{S} \times \mathcal{A} \rightarrow \mathcal{S}$ is the state transition function, $\mathcal{R}$ : $\mathcal{S} \times \mathcal{A} \times \mathcal{S} \rightarrow \mathbb{R}$ is the reward function.
At each timestep, the RL agent takes action $a \in \mathcal{A}$ in current state $s \in \mathcal{S}$ and observes the next state $s^\prime \in \mathcal{S}$ with instant reward $r \in \mathbb{R}$, which forms a quadruple ($s, a, r, s^\prime$) called a \textit{transition}, or an \textit{experience}.
Usually, the objective of RL is to make the agent learn a policy $\pi : \mathcal{S} \rightarrow \mathcal{A}$ that maximizes the cumulative reward $R_c = \mathbb{E}[\sum_{t} r_t]$ when the agent follows it to choose the actions.

When using experience replay, at each timestep, the RL agent interacts with the environment and generates an experience $\mathcal{T}_{new}$ which would be stored into the replay buffer $\mathcal{B}$. When $\mathcal{B}$ is full, some old experiences already in $\mathcal{B}$ will be replaced (or recycled, in our method). Then at a certain frequency of training, the policy (parameters) of RL agent is updated by a batch of experiences $\left\lbrace \mathcal{T}_i \right\rbrace$ sampled from $\mathcal{B}$. The sub-problem we focus on is to learn the sampling mapping $\phi : \mathcal{B} \rightarrow \left\lbrace \mathcal{T}_i \right\rbrace$ and the replacing (and recycling) mapping $\tau : \mathcal{B} \times \mathcal{T}_{new} \rightarrow \mathcal{B}^\prime$, such that the cumulative reward $R_c$ is maximized when the agent samples and replaces experiences based on them.
\begin{figure}[H]
	\centering
	\includegraphics[width=0.75\linewidth]{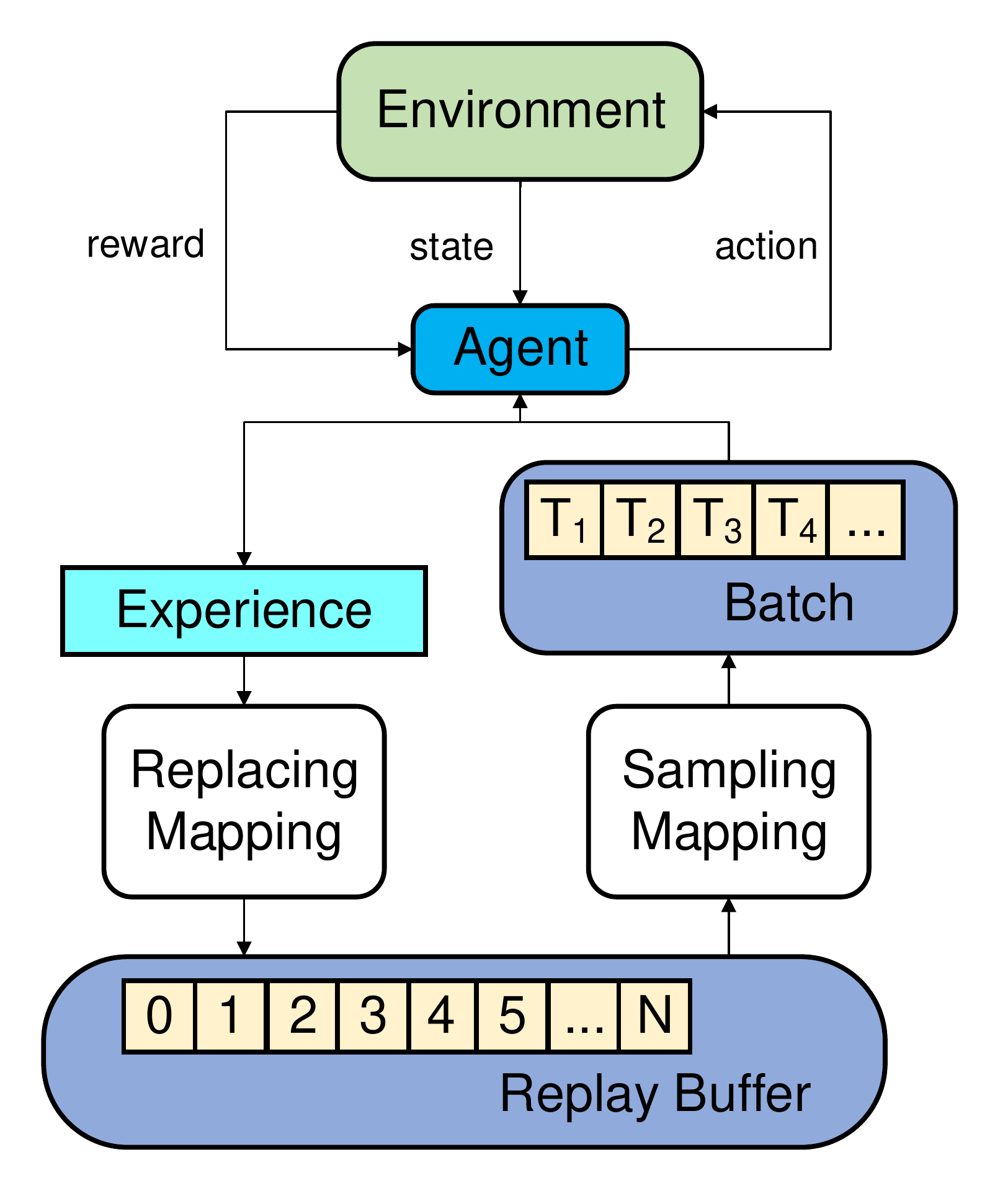}
	\caption{Problem overview}
	\label{fig:probstm}
\end{figure}

\subsection{Prioritized Experience Replay}
In \textit{prioritized experience replay}, TD errors are used to represent the priorities of experiences, which does harm to the diversity of data and produces bias at the same time. To address this issue, \textit{stochastic prioritization} was introduced, where the probability of an experience to be replayed $P(i) = \frac{p_i^\alpha}{\sum_k p_k^\alpha} $ so as to guarantee a non-zero probability for an experience as long as it has a non-zero TD error, where $p_i$ is the priority of $\mathcal{T}_i$, and $\alpha$ is a parameter describing how determining the priorities are in sampling (when $\alpha = 0$, it degenerates to uniform sampling).

As for the detailed formulation of priority, $p_i = |\delta_i| + \epsilon$ for \textit{proportional prioritization}, and $p_i = \frac{1}{rank(i)}$ for \textit{rank-based prioritization}, where $\delta_i$ is the TD error of $\mathcal{T}_i$, $\epsilon$ is a small positive number, and $rank(i)$ is the index of $\mathcal{T}_i$ when sorted by $|\delta_i|$.

Additionally, to anneal the bias caused by non-uniform sampling, importance-sample weights are used, the weight of $\mathcal{T}_i$ for updating the Q-table (Q-network) $w_i = (\frac{1}{N} \cdot \frac{1}{P(i)})^\beta = (N \cdot P(i))^{- \beta}$, where $N$ is the size of replay buffer and $\beta$ is a parameter deciding the ratio of bias-annealing (when $\beta = 1$, the bias is completely settled).

\section{Double Prioritized State Recycled Replay}
When experiences are replayed, the efficiency of training are mainly decided by two things, the quality of the experiences in the replay buffer and the way we choose the experiences to replay. In this paper, the two problems above are addressed separately, and finally an integrated method is designed.

\subsection{Prioritized Sampling}\label{PS}
In sampling, like \textit{prioritized experience replay}, we compute the priorities of experiences based on TD errors and apply \textit{proportional prioritization}. $\mathcal{T}_i$'s priority $p_i = |\delta_i| + \epsilon$, where $\delta_i$ is the TD error of $\mathcal{T}_i$, and $\epsilon$ is a small positive number used to make up the experiences with near-zero TD errors, then its probability of being chosen $PS_i = \frac {p_i^{\alpha(t)}} {\sum_j p_j^{\alpha(t)}}$, where $\alpha(\cdot)$ is the importance-sampling factor that can change over time, indicating how much the probabilities are affected by the priorities.

After the sampling of a batch is completed, we train the Q-network on every experience in the batch with corresponding weight $w_i = \frac {(N \cdot P(i))^{- \beta(t)}} {w_{max}}$, where $N$ is the size of replay buffer, $\beta(\cdot)$ is the bias-annealing factor that can change over time, indicating how strong the bias-annealing is, $t$ is the current timestep, and $w_{max} = \max_{j} (N \cdot P(j))^{- \beta(t)}$ is the max weight among all experiences in replay buffer currently at time $t$.

\subsection{Prioritized Replacing}\label{PR}
Now, let's consider the procedure of replacing old experiences when the replay buffer is full. Briefly speaking, what we want is to find the most likely useless one in the replay buffer and replace it.

Both the original experience replay and prioritized experience replay replace the oldest experience in replay buffer with the latest one when it is full. However, an experience was generated early does not necessarily mean it is useless. Like human beings' intuition, sometimes your first decision without any rational processes is surprisingly good.

Therefore, we propose a method to balance the oldness and usefulness, where first we select some candidates of replacing based on priorities and then find the oldest one among them and choose it as the experience to replace with the newly generated experience.

In detail, $\mathcal{T}_i$'s possibility of being chosen to be a replacing candidates $PR_i = \frac {p_i^{-\gamma(t)}} {\sum_j p_j^{-\gamma(t)}}$, where $p_i$ is the same priority used in section \ref{PS}, and $\gamma(\cdot)$ is the importance-replacing function on time.

When a new experience is inserted, its priority (TD error) would be set to be the max priority currently existing in replay buffer.

\subsection{State Recycling}
In the replacing stage, state recycling would be executed in a certain frequency. To state it simply, what state recycling does is keeping the old state of an experience and generating a new one with the old state and a new action selected by the latest Q-network. To demonstrate it better, we introduce a motivating example which is simple but interesting.

\begin{figure}[H]
	\centering
	\includegraphics[width=0.9\linewidth]{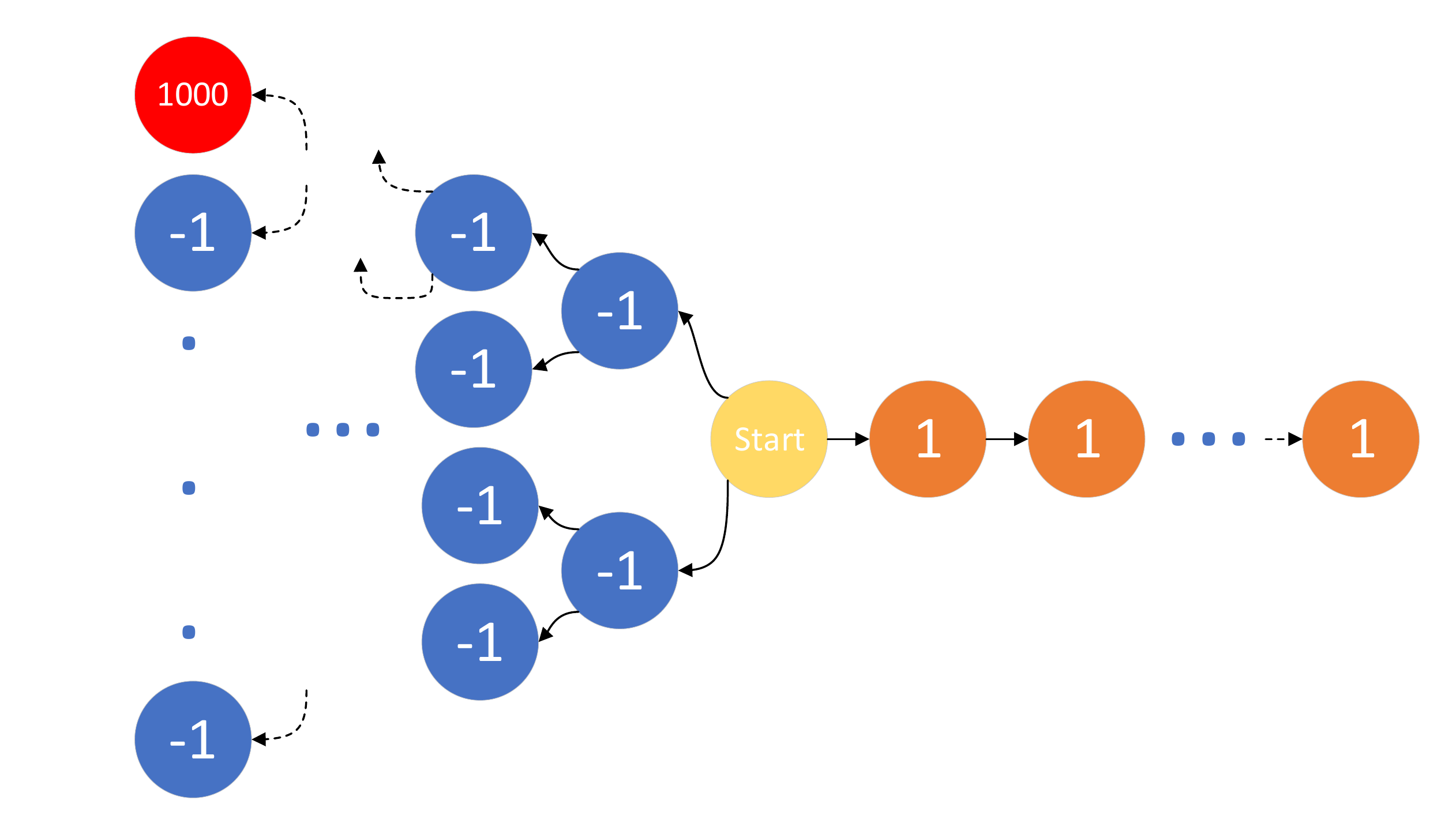}
	\caption{Illustration of the motivating example}
	\label{fig:motivatingex}
\end{figure}

As we can see, once the agent chooses a direction in the first step, it must follow the same direction till the end. In this environment, the agent can easily find a policy which keeps going right with cumulative positive reward. However, as we are omniscient, we know the best policy is find the big treasure in the left part, which can provide the highest reward as long as the depth of it $d \leq 500$. With state recycling, the probability of finding this path is much higher.

In the common case, when it is time to do state recycling in replacing stage, before we replace an old experience already in replay buffer, we choose certain amount of state-recycling candidates according to their priorities. For every experience in the candidates, we keep its old state, and then input it into the latest Q-network to get a new action choice (if it is same as the old one, we will choose a new action randomly). Then we execute this new action for one step in corresponding environment to get a group of new experiences. Finally, we calculate the TD errors of them and find the one with lowest TD error to replace, completing the replacing stage.\\
\hspace*{\fill}

\begin{breakablealgorithm}
	\small
	\caption{DQN with DPSR}
	\renewcommand{\algorithmicrequire}{\textbf{Input: }}
	\noindent \algorithmicrequire minibatch-size $k$, learning rate $\eta$, replay buffer size $N$, exploration function $\epsilon (\cdot)$, importance-sampling function $\alpha (\cdot)$, bias-annealing function $\beta (\cdot)$, importance-replacing function $\gamma (\cdot)$, max priority set flag for state recycling $M$, common replacing candidates size $C_c$, state recycling candidates size $C_r$, target network updating frequency $F_t$, sampling frequency $F_s$, state recycling frequency $F_r$, total timesteps $T$
	\begin{algorithmic}[1]
		\small
		\State Initialize replay buffer $\mathcal{B} = \emptyset$, $\Delta = 0$
		\State Initialize the parameters of Q-network $\theta$ randomly, and initialize $\theta_{target} = \theta$
		\State Observe $s_0$ and choose $a_0 \sim \pi_\theta(s_0)$ with $\epsilon(0)$ possibility to choose an action randomly
		\For {$t = 1$ \textbf{to} $T$}
		\State Observe $s_t, r_t$
		\State Assemble experience $(s_{t-1}, a_{t-1}, r_t, s_t)$ with $p_t = \max_{i<t} p_i$ 
		\If {$|\mathcal{B}| < N$}
		\State Add the new experience at the end of the queue
		\ElsIf {$t \equiv 0 \bmod F_r$} 
		\For {$i = 1$ \textbf{to} $C_r$}
		\State Sample experience $(s_{t_i-1}, a_{t_i-1}, r_{t_i}, s_{t_i}) \sim PR(t_i) = p_{t_i}^{-\gamma (t)} / \sum_j p_j^{-\gamma (t)}$
		\State Input $s_{t_i-1}$ into the latest Q-network with $\theta$ and get its choice of action $a^\prime_{t_i-1}$
		\If {$a^\prime_{t_i-1} = a_{t_i-1}$}
		\State Randomly choose an available action $\tilde{a}_{t_i-1} \neq a_{t_i-1}$, $a^\prime_{t_i-1} \leftarrow \tilde{a}_{t_i-1} $
		\EndIf
		\State Observe $s^\prime_{t_i}, r^\prime_{t_i}$
		\If {$M = True$ }
		\State $p_{t_i} = \max_{i<t} p_i$
		\Else
		\State Compute TD error $\delta_{t_i} = r^\prime_{t_i} + Q_{target}(s^\prime_{t_i}, \arg \max_a Q(s^\prime_{t_i}, a)) - Q(s_{t_i-1}, a^\prime_{t_i-1})$
		\State Update the priority $p_{t_i} \sim |\delta_{t_i}|$
		\EndIf
		\State Assemble experience $(s_{t_i-1}, a^\prime_{t_i-1}, r^\prime_{t_i}, s^\prime_{t_i})$ with new $p_{t_i}$ obtained above
		\EndFor
		\State Let $i^* = \arg \min_{1\leq j \leq C_r} p_{t_j}$, and replace the corresponding experience $\mathcal{B}_{t_{i^*}}$ with the new one
		\Else 
		\For {$i = 1$ \textbf{to} $C_c$}
		\State Sample experience $(s_{t_i-1}, a_{t_i-1}, r_{t_i}, s_{t_i}) \sim PR(t_i) = p_{t_i}^{-\gamma (t)} / \sum_j p_j^{-\gamma (t)}$
		\EndFor
		\State Let $i^* = \arg \min_{1\leq j \leq C_c} t_j$, and replace the corresponding experience $\mathcal{B}_{t_{i^*}}$ with the new one
		\EndIf
		\If {$t \equiv 0 \bmod F_s$} 
		\For {$i = 1$ \textbf{to} $k$}
		\State Sample experience $(s_{i-1}, a_{i-1}, r_i, s_i) \sim P(i) = p_i^{\alpha (t)} / \sum_j p_j^{\alpha (t)}$
		\State Compute importance-sampling weight $w_i = (N \cdot P(i))^{- \beta (t)}$, then let $w_i \leftarrow w_i / \max_{j} w_j$
		\State Compute TD error $\delta_i = r_i + Q_{target}(s_i, \arg \max_a Q(s_i, a)) - Q(s_{i-1}, a_{ai-1})$
		\State Update the priority of experience $p_i \sim |\delta_i|$
		\State Accumulate weight-change $\Delta \leftarrow \Delta + w_i \cdot \delta_i \cdot \nabla_\theta Q(s_{i-1}, a_{i-1})$  
		\EndFor
		\State Update weight $\theta \leftarrow \theta + \eta \cdot \Delta$, $\Delta \leftarrow 0$
		\If {$t \equiv 0 \bmod F_t$}
		\State $\theta_{target} \leftarrow \theta $
		\EndIf
		\EndIf
		\State Choose action $a_t \sim \pi_{\theta}(s_t)$ with $\epsilon(t)$ possibility to choose an action randomly
		\EndFor
	\end{algorithmic}
\end{breakablealgorithm}

\section{Experiments Results}
Now, we are going to show the performance of our method in realistic problem domains. We completed the implementation based on \textit{OpenAI Gym} [\citealp{1606.01540}] platform and compared our method with the original experience replay and prioritized experience replay which are provided as baseline methods in the platform.

To ensure the fairness, we use the identical deep neural network architecture and most parameters of learning algorithm such as mini-batch size, learning rate, replay buffer size, exploration policy, importance-sampling and bias-annealing factors, sampling frequency, total timesteps and so on, which are shown comprehensively in Table \ref{tab:param}.

\begin{figure}
	\centering
	\includegraphics[width=\linewidth]{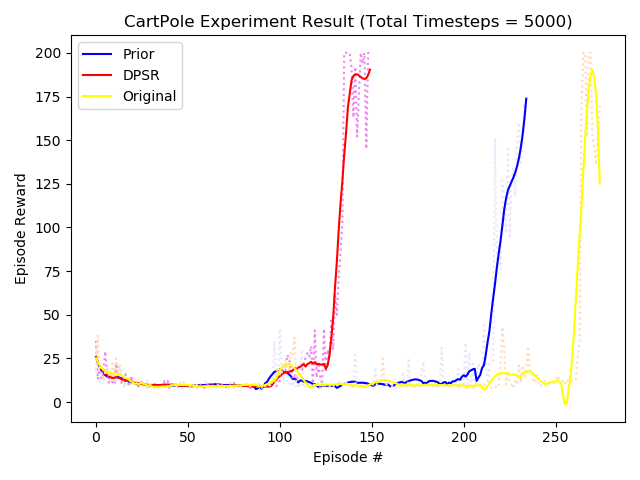}
	\caption{Experiment results in cartpole}
	\label{fig:cartpole}
\end{figure}

First, we tested our method in a simple environment \textit{cartpole}, which is a classic control problem. As shown in Figure \ref{fig:cartpole}, our method learns the optimal policy much more efficiently than both the original method and prioritized experience replay.

Then we completed experiments on more complicated cases, Atari game environments. We train RL agents with the original experience replay, prioritized experience replay, and DPSR experience replay separately under basically same common parameter sets and get the test results shown in Table \ref{tab:atari}. In all the 24 games, our method wins 23 "gold medals" and 1 "silver medal". The average and median of the performance improvement are 161.1\% and 87.0\% compared to the original method, while the numbers are 137.1\% and 92.1\% compared to prioritized experience replay (\textit{JourneyEscape} and \textit{Zaxxon} are excluded as both baseline methods get non-positive scores in these two games). Besides, we'd like to note that original experience replay outperforms prioritized experience replay in 14 games, and even wins one "gold medal" in \textit{SpaceInvaders}, which is a little surprising. 

\begin{table}
  \centering
	\caption{Experiment results in Atari games}
	\begin{threeparttable}
  \begin{tabular}{llll}
    \toprule
    Game name  		& Original	& Prioritized	& DPSR\_best\tnote{*}		\\
    \midrule
    AirRaid			& 545.0		& 602.5			& \textbf{4182.5}	\\
    Alien			& 853.0		& 917.0			& \textbf{1824.0}	\\
    Amidar			& 155.0		& 145.7			& \textbf{294.6}	\\
    Assault 		& 882.5		& 638.9 		& \textbf{990.0}	\\
	Asterix 		& 2055.0	& 1435.0		& \textbf{2430.0}	\\
	BeamRider 		& 2101.8	& 2442.0		& \textbf{2558.8}	\\
	Bowling 		& 29.0		& 24.2 			& \textbf{60.5}		\\
	Breakout 		& 87.0		& 136.4		 	& \textbf{281.4}	\\
	Carnival 		& 3286.0	& 2006.0 		& \textbf{3981.0}	\\
	Enduro 			& 612.3 	& 497.3 		& \textbf{1025.1}	\\
	Freeway 		& 30.2 		& 29.5 			& \textbf{32.2}		\\
	Frostbite 		& 229.0 	& 998.0 		& \textbf{2186.0}	\\
	Hero 			& 2891.5 	& 2585.0 		& \textbf{11060.0}	\\
	JourneyEscape 	& -4150.0 	& -3350.0 		& \textbf{4440.0}	\\
	Krull 			& 4736.7 	& 5541.3 		& \textbf{9406.3}	\\
	KungFuMaster 	& 14900.0 	& 19820.0 		& \textbf{28300.0}	\\
	MsPacman 		& 1682.0 	& 1666.0 		& \textbf{3095.0}	\\
	Phoenix 		& 3668.0 	& 2787.0 		& \textbf{4380.0}	\\
	Qbert 			& 1980.0 	& 885.0 		& \textbf{4525.0}	\\
	Riverraid 		& 5334.0 	& 4596.0 		& \textbf{5792.0}	\\
	SpaceInvaders   & \textbf{610.0}& 316.5 	& 580.5				\\
	StarGunner 		& 2050.0 	& 1490.0 		& \textbf{2490.0}	\\
	VideoPinball 	& 7313.0 	& 12025.1 		& \textbf{51993.2}	\\
	Zaxxon 			& 0.0 		& 0.0 			& \textbf{3640.0}	\\
    \bottomrule
  \end{tabular}
	\begin{tablenotes}
		\footnotesize
		\item[*] The parameter sets achieving the best performance in each game respectively
	\end{tablenotes}
	\end{threeparttable}
  \label{tab:atari}
\end{table}

\section{Discussion}
To get better performance, we tried many different sets of hyperparameters and found that the method with only prioritized sampling and prioritized replacing (state recycling disabled) can have quite bad performance in some games. We guess this phenomenon may be caused by the fact that both prioritized sampling and prioritized replacing would introduce bias, therefore when we only use them, the double-bias can cause some negative effect on performance. We use the same bias-annealing factor for prioritized experience replay and DPSR experience replay in order to maintain the fairness but we can rationally guess that DPSR experience replay may have better performance with stronger bias-annealing techniques because of the reason mentioned above.

When we decide state recycling frequency and two replacing candidates sizes, target network updating frequency and the computational cost for state recycling should be taken into consideration. That is why we keep these parameters in a small range.

Another important trick we use in state recycling is to ensure that experiences after state recycling contain different actions with the previous one, otherwise the process can be totally waste of computation and time, as shown in our previous experiments.

Currently, we do state recycling by saving the full state of the environment which can be further optimized by estimating the whole environment with only part of the state saved. We are still trying to find a proper way to do this.

\section{Conclusion}
In this paper, we proposed double-prioritized state-recycled (DPSR) experience replay, a method that can make RL agents learn more efficiently. We compared our method with original experience replay and prioritized experience replay in some simple environments and Atari game environments, achieving state-of-the-art results.

\section*{Acknowlement}
  This research has been supported in part by the ICT R\&D program of MSIP/IITP 2016-0-00563 [Research on Adaptive Machine Learning Technology Development for Intelligent Autonomous Digital Companion], and MSIP/IITP 2019-0-01396 [Development of framework for analyzing, detecting, mitigating of bias in AI model and training data].

\nocite{*}
\bibliographystyle{ieeetr}
\bibliography{references}

\titleformat{\section}[block]{\large \bfseries}{{\thesection}. }{0pt}{}[]
\appendix
\section{Implementation Details}\label{ID}
We completed our implementation and experiments based on the \textit{deepq} module on \textit{OpenAI Gym} platform. The environments of Atari game we used are in NoFrameskip-v4 version (e.g. The environment name of the SpaceInvaders game is SpaceInvadersNoFrameskip-v4). There are some common hyperparameters (some may not used in original method and prioritized method) and also some inconstant hyperparameters for our method. See Table \ref{tab:param} for more details.

\begin{table}[H]
	\centering
	\caption{Hyperparameters settings}
	\begin{tabular}{ll}
		\toprule
		Hyperparameter & Value (Range of values) \\
		\midrule
		$k$ 			& $32$ 														\\
		$\eta$ 			& $0.0005$ 													\\
		$N$ 			& $50000$ 													\\
		$\epsilon(t)$ 	& $max(1 - 9.8 t / T, 0.02)$ 								\\
		$\alpha(t)$ 	& $0.6$ 													\\
		$\beta(t)$ 		& $0.4 + 0.6 t / T$ 										\\
		$\gamma(t)$ 	& $0.1, 0.2, 0.3, 0.4, 0.5, 0.6$							\\
		$C_c$ 			& $128, 256$												\\
		$C_r$ 			& $8, 16, 32, 64$											\\
		$F_t$ 			& $500$ 													\\
		$F_s$ 			& $1$ 														\\
		$F_r$ 			& $10000, 20000$ 											\\
		$T$   			& $1000000$\\		
		\bottomrule
	\end{tabular}
	\label{tab:param}
\end{table}

Besides, as shown in Table \ref{tab:bestparam}, for different game environments, the best performance of our method may be achieved by different setting of parameters. But we can still find some settings of parameters which have relatively good performance in most games as shown in Table \ref{tab:goodparams}.

\begin{table*}
	\caption{Best parameters in different games}
	\centering
	\begin{threeparttable}
	\begin{tabular}{ll}
		\toprule
		Game name 		& Parameters achieving the top 3 scores ($\gamma, C_c, F_r, C_r$) \\
		\midrule
		AirRaid 		& (.5, 128, 10k, 16), (.2, 128, 20k, 8),  (.2, 128, 10k, 8) \\
		Alien 			& (.5, 256, 20k, 16), (.6, 256, 0, 0)\tnote{*}, 	  (.2, 256, 20k, 64) \\
		Amidar 			& (.1, 128, 10k, 8), 	(.3, 256, 0, 0), 	  (.1, 128, 10k, 16) \\
		Assault 		& (.6, 256, 10k, 64), (.1, 128, 20k, 8),  (.5, 128, 10k, 8) \\
		Asterix 		& (.2, 128, 10k, 16), (.1, 256, 20k, 64), (.3, 128, 20k, 8) \\
		BeamRider 		& (.4, 256, 10k, 8),  (.1, 128, 0, 0),      (.3, 128, 10k, 8) \\
		Bowling 		& (.6, 256, 10k, 16), (.6, 256, 20k, 64), (.6, 128, 20k, 8) \\
		Breakout 		& (.1, 128, 10k, 64), (.3, 128, 10k, 8),  (.2, 128, 20k, 32) \\
		Carnival 		& (.1, 128, 10k, 32), (.1, 256, 10k, 16), (.6, 128, 20k, 8) \\
		Enduro 			& (.6, 256, 10k, 64), (.6, 256, 10k, 64), (.6, 128, 10k, 16) \\
		Freeway 		& (.1, 256, 10k, 8), 	(.3, 128, 20k, 16), (.3, 128, 20k, 64) \\
		Frostbite 		& (.1, 256, 20k, 8), 	(.4, 256, 20k, 16), (.6, 128, 10k, 16) \\
		Hero 			& (.5, 128, 10k, 32), (.6, 256, 10k, 64), (.4, 256, 10k, 64) \\
		JourneyEscape 	& (.6, 256, 10k, 64), (.3, 128, 10k, 8),  (.2, 128, 20k, 64) \\
		Krull 			& (.5, 256, 20k, 16), (.1, 256, 20k, 64), (.2, 128, 20k, 64) \\
		KungFuMaster 	& (.6, 128, 20k, 32), (.4, 128, 20k, 16), (.3, 256, 20k, 64) \\
		MsPacman 		& (.3, 128, 10k, 32), (.1, 128, 10k, 32), (.4, 256, 20k, 8) \\
		Phoenix 		& (.2, 256, 10k, 16), (.3, 128, 10k, 8),  (.6, 256, 20k, 16) \\
		Qbert 			& (.6, 256, 10k, 32), (.5, 128, 10k, 16), (.6, 128, 10k, 16) \\
		Riverraid 		& (.5, 256, 10k, 32), (.1, 256, 20k, 64), (.6, 128, 10k, 32) \\
		SpaceInvaders 	& (.3, 256, 10k, 8), 	(.1, 128, 10k, 16), (.1, 256, 20k, 64) \\
		StarGunner 		& (.3, 128, 20k, 16), (.2, 128, 10k, 16), (.3, 128, 10k, 8) \\
		VideoPinball 	& (.1, 128, 0, 0), 		(.1, 128, 20k, 16), (.4, 256, 20k, 8) \\
		Zaxxon 			& (.1, 256, 20k, 64), (.3, 128, 10k, 64), (.1, 256, 20k, 32) \\
		\bottomrule
	\end{tabular}
	\begin{tablenotes}
		\footnotesize
		\item[*] $F_r = C_r = 0$ means state recycling is disabled
	\end{tablenotes}

	\end{threeparttable}
	\label{tab:bestparam}
\end{table*}

\begin{table*}
	\caption{Experiment results of some parameter sets}
	\centering
	\begin{threeparttable}
  	\begin{tabular}{llllllll}
	\toprule 
	Game name  		& Original	& Prioritized	& DPSR0\tnote{*} & DPSR1 & DPSR2 & DPSR3 & DPSR4\\
	\midrule
	AirRaid & 545.0 & 602.5 & 2892.5 & 1285.0 & 632.5 & \textbf{3175.0} & 1150.0 \\
	Alien & 853.0 & 917.0 & 995.0 & 673.0 & 942.0 & \textbf{1111.0} & 750.0 \\
	Amidar & 155.0 & 145.7 & 148.1 & \textbf{221.8} & 155.5 & 164.3 & 136.4 \\
	Assault & \textbf{882.5} & 638.9 & 651.8 & 714.3 & 704.9 & 675.7 & 676.9 \\
	Asterix & 2055.0 & 1435.0 & 1600.0 & \textbf{2145.0} & 1635.0 & 1180.0 & 1635.0 \\
	BeamRider & 2101.8 & \textbf{2442.0} & 1525.6 & 1899.2 & 1086.0 & 2317.6 & 1021.6 \\
	Bowling & 29.0 & 24.2 & \textbf{44.4} & 9.6 & 2.0 & 29.1 & 31.8 \\
	Breakout & 87.0 & 136.4 & 134.5 & 80.6 & 91.5 & 129.6 & \textbf{145.8} \\
	Carnival & 3286.0 & 2006.0 & \textbf{3847.0} & 2704.0 & 3356.0 & 860.0 & 2817.0 \\
	Enduro & 612.3 & 497.3 & 795.4 & 503.6 & 725.9 & 771.4 & \textbf{919.2} \\
	Freeway & 30.2 & 29.5 & 30.7 & \textbf{31.7} & 30.5 & 20.6 & \textbf{31.7} \\
	Frostbite & 229.0 & 998.0 & 382.0 & 664.0 & 217.0 & 1502.0 & \textbf{1505.0} \\
	Hero & 2891.5 & 2585.0 & 5937.0 & 2936.0 & 2838.5 & \textbf{6534.0} & 3623.0 \\
	JourneyEscape & -4150.0 & -3350.0 & -6520.0 & -2130.0 & -4290.0 & \textbf{140.0} & -3040.0 \\
	Krull & 4736.7 & 5541.3 & 5864.4 & \textbf{6418.8} & 6001.2 & 6279.3 & 6248.7 \\
	KungFuMaster & 14900.0 & 19820.0 & 14500.0 & 16020.0 & 13990.0 & 13240.0 & \textbf{20490.0} \\
	MsPacman & 1682.0 & 1666.0 & \textbf{1926.0} & 1913.0 & 1768.0 & 1742.0 & 1757.0 \\
	Phoenix & \textbf{3668.0} & 2787.0 & 3234.0 & 3122.0 & 3342.0 & 3127.0 & 1991.0 \\
	Qbert & 1980.0 & 885.0 & \textbf{2412.5} & 2167.5 & 2002.5 & 2120.0 & 565.0 \\
	Riverraid & 5334.0 & 4596.0 & 5073.0 & \textbf{5349.0} & 4949.0 & 3175.0 & 3194.0 \\
	SpaceInvaders & \textbf{610.0} & 316.5 & 510.0 & 579.5 & 490.0 & 406.0 & 545.0 \\
	StarGunner & 2050.0 & 1490.0 & 870.0 & 1680.0 & 1440.0 & \textbf{2180.0} & 1890.0 \\
	VideoPinball & 7313.0 & 12025.1 & \textbf{30337.0} & 21596.9 & 15689.0 & 16650.0 & 9483.6 \\
	Zaxxon & 0.0 & 0.0 & 0.0 & 0.0 & \textbf{1270.0} & 0.0 & 0.0 \\
	\bottomrule
	\end{tabular}
	\begin{tablenotes}
		\footnotesize
		\item[*]$\gamma, C_c, F_r, C_r = (.6, 128, 20k, 8), (.1, 128, 10k, 16), (.2, 256, 20k, 8), (.3, 128, 10k, 8), (.3, 256, 10k, 16)$\\
	\end{tablenotes}

\end{threeparttable}
\label{tab:goodparams}
\end{table*}
	
\end{document}